\documentclass[letterpaper, 10 pt, conference]{ieeeconf}  
\usepackage{mathtools}
\usepackage{url}
\usepackage{siunitx}
\usepackage[dvipsnames]{xcolor}
\usepackage{amsmath,amssymb,relsize}
\usepackage{multirow}
\usepackage[ruled,vlined,linesnumbered]{algorithm2e}
\usepackage[export]{adjustbox}
\usepackage{dblfloatfix}
\usepackage{placeins}

\newcommand{\rev}[1]{\textcolor{black}{#1}}

\DeclareMathOperator{\MMD}{MMD}

\newcommand{\realtdwidth}{0.08}
\newcommand{\realddwidth}{0.12}
\newcommand{\arttdwidth}{0.05}

\IEEEoverridecommandlockouts                              

\title{\LARGE \bf
DLOFTBs -- Fast Tracking of Deformable Linear Objects with B-splines}

\author{Piotr Kicki$^{1}$, Amadeusz Szymko$^{1}$ and Krzysztof Walas$^{1}$
\thanks{This work is supported by the European Union's Horizon 2020 Research and Innovation Programme under grant agreement No 870133, REMODEL.}
\thanks{$^{1}$ Institute of Robotics and Machine Intelligence, Poznan University of Technology, Poznan, Poland; {e-mail: \tt\small \{name.surname\}@put.poznan.pl}}%
}

\begin{document}

\maketitle
\thispagestyle{empty}
\pagestyle{empty}

\begin{abstract}

While manipulating rigid objects is an extensively explored research topic, deformable linear object (DLO) manipulation seems significantly underdeveloped. A potential reason for this is the inherent difficulty in describing and observing the state of the DLO as its geometry changes during manipulation.
This paper proposes an algorithm for fast-tracking the shape of a DLO based on the masked image. Having no prior knowledge about the tracked object, the proposed method finds a reliable representation of the shape of the tracked object within tens of milliseconds. 
This algorithm's main idea is to first skeletonize the DLO mask image, walk through the parts of the DLO skeleton, arrange the segments into an ordered path, and finally fit a B-spline into it.
Experiments show that our solution outperforms the State-of-the-Art approaches in DLO's shape reconstruction accuracy and algorithm running time and can handle challenging scenarios such as severe occlusions, self-intersections, and multiple DLOs in a single image.

\end{abstract}
\section{Introduction}
Deformable Linear Objects (DLOs) are a class of objects that are characterized by two main features: deformability, which refers to the fact that the object is not a rigid body and its geometry can change, and linearity, which stands for the fact that the object is elongated and the ratio of its length to its width is substantial~\cite{survey}. Such objects are ubiquitous both in everyday life and in industry, where one can find ropes, cables, pipes, sutures, etc. 
While the manipulation of rigid bodies is already solved for a wide range of objects~\cite{trends_and_challenges}, manipulating DLOs is still unsolved even for everyday objects such as cables and hoses. Due to the ubiquity of the DLOs, manipulating them poses a complex and vital challenge, which has been in the scope of researchers for over three decades~\cite{oldest_dlo_manip}. The interest in this topic has grown over the last few years, as the automatic wiring harness assembly is crucial for car manufacturers~\cite{wiring_harness_assembly}, as well as automatic completion of surgical sutures, which could help surgeons~\cite{surgical_sutures}.
\rev{To perform such tasks autonomously, robotic manipulators need to perceive the configuration of the manipulated object, as this is crucial for calculating the adequate control signal.}
\rev{Achieving this requires accurate and fast DLO tracking abilities.}
However, state-of-the-art DLO tracking algorithms are relatively slow and do not meet the time requirements of the control systems. In addition, they can not properly handle manipulation sequences that contain occlusions~\cite{random_matrices} and self-intersections~\cite{cdcpd2}, or make many assumptions about the model of the tracked object~\cite{apetit_siciliano, tomizuka_gmm}.

\begin{figure}[th]
    \centering
    \includegraphics[width=0.98\linewidth]{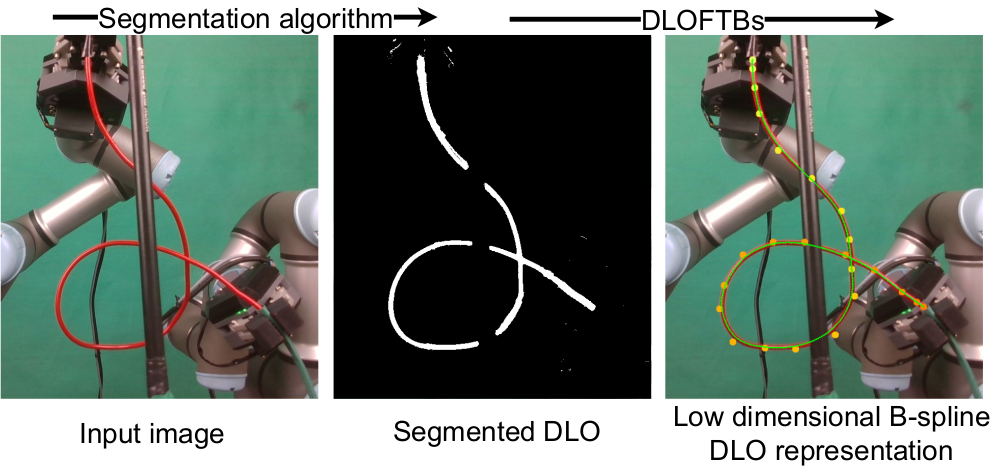}
    \vspace{-0.4cm}
    \caption{Using the proposed DLO tracking algorithm, one can transform the DLO mask into a low-dimensional B-spline representation in less than \SI{40}{\milli\second}.}
    \label{fig:scheme}
    \vspace{-0.4cm}
\end{figure}

This paper proposes a fast, non-iterative method for estimating a DLO's shape using a walk through object's mask and B-spline regression. The proposed algorithm takes as an input the mask of the DLO and returns a sequence of control points of the B-spline curve that approximates the shape of the tracked DLO. Our solution can deterministically identify the shape of a DLO on the HD image within $\SI{40}{\milli\second}$ while handling non-trivial scenarios, like occlusions, self-intersections, and multiple DLOs in the scene. The general scheme of the proposed approach is presented in Figure~\ref{fig:scheme}.

The main contribution of this work is twofold:
\begin{itemize}
    \item a novel deterministic fast DLO tracking algorithm, which can handle occlusions, self-intersections, and multiple DLOs in the scene while requiring no prior knowledge about the tracked DLO and is faster and more accurate than State-of-the-Art solutions for assumed quality of the output shape,
    \item dataset of real and artificial 2D and 3D videos and images of several different DLOs, on which we performed a verification of the proposed method and which we share with the community for objective performance evaluation and to encourage the development of real-time DLO tracking\footnote{\url{https://github.com/PPI-PUT/cable_observer/tree/master}}.
\end{itemize}

\section{Related Work}


\subsection{DLO representation}
In the literature, there are several ways to represent the geometric shape of the DLO. The most straightforward one is representing it as a sequence of points~\cite{RL_DLO_manipulation, fiducial}. However, more complex models are usually necessary for accurate cable modeling and tracking, like a B-spline model with multiple chained random matrices, proposed in~\cite{random_matrices}. A similar approach, but using Bezier curves and rectangle chains, was proposed in~\cite{bezier}, while in~\cite{nurbs} NURBS curves were used.
In our research, we use a B-spline representation (similar to the one used in~\cite{bezier, nurbs}) as it is flexible and enables one to accurately track the shape of a generic DLO while being compact, relatively easy, and cheap to work with. Using this representation, one can build more complex models, which consider the kinematics and dynamics of the DLO~\cite{gianluca_ICPS, cosserat_rod}.

\subsection{DLO tracking}
DLO tracking requires transforming the data gathered with sensors into the chosen representation. 
While there are attempts to use data from tactile sensors~\cite{FEM}, the most successful way to perceive the DLO shape is to use vision and depth sensors. One of the most straightforward approaches to DLO shape tracking is to use the fiducial markers located along the DLO, and track them~\cite{fiducial} or use them to estimate the shape of a DLO~\cite{bretl}. A similar approach was presented in~\cite{tangled}, where colors denote consecutive rope segments.
The most common approach is to create a model of the DLO and use images or point clouds as measurements to modify its parameters and track the object deformation iteratively. One of the examples of this approach is the modified expectation-maximization algorithm (EM), proposed in~\cite{abbeel_probabilistic}, which is used to update the predefined DLO model based on the registered deformations and simulation in the physics engine. Similarly, in~\cite{apetit_siciliano}, the FEM methods were used to track the deformation of the predefined model. Whereas, in~\cite{tomizuka_gmm}, a Structure Preserved Registration algorithm with the object represented as a Mixture of Gaussians was used. Authors of~\cite{bezier} performed DLO tracking using the Recursive Bayesian Estimator on the Spatial Distribution Model, built with the Bezier curve and the chain of rectangles.
Due to the iterative and often probabilistic character of the model updates, these methods usually have problems tracking rapidly deforming objects and require an appropriate model and accurate initialization. 
To mitigate the slow initialization problem, authors of~\cite{random_matrices} used the Euclidean minimum spanning tree and the Breadth-first search method to speed up initialization. However, obtaining the DLO shape estimate still takes hundreds of milliseconds.
A much faster EM-based tracking approach, which utilizes a coherent point drift method extended with some geometric-based regularization and physically and geometrically inspired constraints, was presented in~\cite{cdcpd2}.
The instance segmentation method for multiple DLOs, which also can serve for tracking, was initially proposed in~\cite{ariadne} and extended using Deep Learning solutions in~\cite{ariadne_plus} and~\cite{fastdlo}.

The solution presented in this paper utilizes a similar idea to the one presented in~\cite{ariadne_plus, fastdlo}. However, using skeletons instead of the super-pixel graphs reduces the computational complexity~\cite{ariadne_plus}, and the lack of Deep Learning in our solution facilitates better generalization without sacrificing performance~\cite{ariadne_plus, fastdlo}. In our work, we do not try to model the DLO, but instead, quickly provide a compact representation of the DLO state\rev{, that is consistent and is trackable between frames}. Thus, the proposed method can aid the existing model-based methods with fast and accurate structured measurements of the system state.

\section{DLO Tracking}
\subsection{Problem Formulation}
\label{sec:problem_formulation}
The problem considered in this paper is to track the DLO on the video sequence. By tracking, we understand transforming consecutive video frames of the DLO's binary mask, obtained from the selected segmentation algorithm, into a 1D curve resembling the object's shape\rev{, which representation should be consistent between frames}. 
In this paper, we will not consider the image segmentation problem similarly to~\cite{random_matrices, cdcpd2, abbeel_probabilistic}. But instead, we will focus on shape tracking only, with the assumption that for homogeneously colored cables, the mask is given by any color-based segmentation algorithm, or in more challenging scenarios, state-of-the-art deep learning method~\cite{ariadne_plus} is used.




\subsection{Proposed Method}

\begin{figure*}[t]
    \centering
    \includegraphics[width=\linewidth]{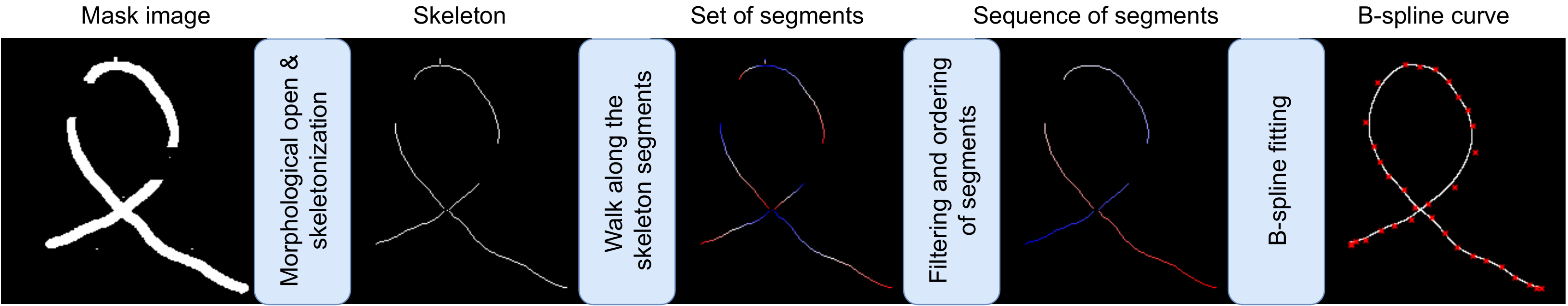}
    \caption{The general scheme of DLOFTBs. The mask image is transformed through several processing stages to obtain a B-spline representation of its shape.}
    \label{fig:solution}
    \vspace{-0.35cm}
\end{figure*}

In this section, we introduce our proposed novel approach to fast tracking of DLO, called DLOFTBs, which by using the walks on the DLO mask's skeleton, enables rapid fitting of the B-spline curve into the masked image of the DLO. The general scheme of the proposed algorithm is presented in Figure~\ref{fig:solution}. To transform the mask image into a B-spline curve, 4 main processing steps are made: (i) morphological open \& skeletonization, (ii) walk along the skeleton segments, (iii) filtering and ordering of segments, and finally, (iv) B-spline fitting. In the following subsections, we will describe each of these steps in detail. 

\subsubsection{Morphological open \& skeletonization}
\label{sec:skel}
The first operation we perform on the mask of the DLO is a $3 \times 3$ (the smallest possible) morphological open. We used it to remove some false positive pixels, which are common because of imperfect segmentation. After that, one of the essential steps in mask processing is performed -- skeletonization~\cite{skeleton_used}.
This operation takes the mask image as an input and creates its skeleton, i.e., a thin version of the mask, which lies in the geometric centers of the DLO segments, preserves its topology, and reduces its width to a single pixel.
This significantly reduces the amount of information about the pixels representing the DLO while preserving the crucial information encoded in central pixels along the DLO. Moreover, using the skeleton, one can easily find the crucial parts of the DLO mask, such as segment endpoints -- pixels with only one neighbor or branching -- pixels with more than two neighbors. While the segment endpoints will constitute starting points for the walks on segments, the branching points are crucial while performing a walk, as they require the walker to choose one of several possible paths. 
To avoid this inconvenience, we propose removing branching points and postponing the decision to make connections between segments for further processing. By doing so, we can simplify the segment walk algorithm considerably.


\subsubsection{Walk algorithm}
\label{sec:walk}
Having the skeleton prepared and segment endpoints determined, we can perform a walk along each segment. To do so, we start from a random segment endpoint and go pixel by pixel till the end of the segment, collecting the subsequent pixel coordinates. Such traversing is always possible and unambiguous, as we removed all pixels with more than two neighbors in the previous step.
After each walk, we remove two points from the set of endpoints that were the segment's beginning and end.
Next, we draw another segment endpoint and perform a walk, which repeats until the end of the segment endpoints. As a result, we obtain a set of paths i.e., ordered lists of pixels representing all segments.

    

\subsubsection{Filtering and ordering of segments}
\label{sec:order}
In this step, we first filter out segments shorter than $p$ pixels, which are likely to represent some artifacts of the mask or resulting from the skeletonization procedure. While this approach may also result in removing a short part of the actual DLO, it will not affect the resultant path significantly, as it will be treated as occlusion and handled by our algorithm at the next stage.

In order to fit a B-spline effectively into a set of segments, we need to order them. As a result of the previous processing step, we have an unordered set of ordered lists of pixels. To order them, we need to find segment endpoint pairs that will most likely connect to each other. While there are many possible criteria and algorithms for deciding about connections, we decided to use a criterion that takes into account both the distance and orientation of the endpoints and is defined by
\begin{equation}
\label{eq:J}
    J = m J_d + (1 - m) J_o,
\end{equation}
where $J_d$ is a euclidean distance between segment endpoints, $J_o$ is a criterion related to the mutual orientation of the segment endpoints, and $m \in [0; 1]$ is a linear mixing factor. While the definition of $J_d$ is rather straightforward, the exact formula of the $J_o$ is given by
\begin{equation}
    J_d = |\pi - \phi_1 - \phi_2|,
\end{equation}
where $\phi_1, \phi_2$ are approximated orientations of the segment endpoints.

Using the criterion $J$~\eqref{eq:J}, one can decide about the pairs of the segment endpoints. The most accurate solution would be to check for all possible pairing schemes and find the one with the lowest $J$. \rev{However, it is also the most computationally expensive one, as it requires checking even $(2s - 1)(2s - 3)\ldots 1$ pairings, where $s$ is the number of segments. To limit the computational burden, we decided to use a potentially less accurate but much faster approach -- a greedy one. Thus, we need to choose at most $s-1$ connections out of $s(2s-1)$ pairs of endpoints, considering the already taken endpoints. Note that we don't want to make highly improbable connections. Therefore, we introduce a threshold $J_{th}$ and consider only the connections for which criterion $J < J_{th}$. This enables us to track a single cable robustly and detect and track multiple DLOs at once. Finally, all detected sequences of segments are separately passed to the B-spline fitting phase, described in the next point. B-spline curves provide a compact representation that can be used to identify DLO instances based on the representation continuity between frames.}

\subsubsection{B-spline fitting}
\label{sec:fit}
To fit the B-spline to the sequence of segments, we need an argument for the B-spline, i.e., the vector $t$ of the relative position of the pixels on the curve we want to define. To do so, we calculate the distance along the segments and euclidean distances between segments and concatenate them into a single vector, the cumulative sum of which serves as the B-spline argument $t$.
Using the Euclidean distance between segments, we introduce an estimate of the distance along the DLO (we do not have access to the true one, as the parts of a DLO are occluded). This procedure prevents sudden changes in the pixel's positions in terms of the B-spline argument $t$.

Moreover, we need to define the number and positions of knots. In the proposed solution, we defined knots as a sequence of $k$ equidistant, in terms of the element number, elements of the vector $t$, as it will ensure that the Schonenberg-Whitney conditions~\cite{Schoenberg1953} are met.

\rev{Finally, one can fit two dimensional B-spline using the prepared argument vector $t$, knots, and $x$ and $y$ coordinates of the ordered pixels.} 
We used cubic splines, as higher continuity is unnecessary for the considered problem. 

\subsubsection{3D data}
Even though the proposed DLO tracking algorithm is meant to work on images, it can be easily extended to work for the 3D data obtained from the RGBD sensor. 
In this case, we deal with the mask in the same way as for the 2D case till the moment of the B-spline fitting. Given a sequence of segments in the 2D space, we augment it with the corresponding depth coordinates and then perform the B-spline fitting. \rev{Thus, we obtain three dimensional B-spline representing the shape of the cable with respect to the curve length estimate.}

\section{Experiments}
To perform \rev{all} experiments, we used a single core of the Intel Core i7-9750H CPU and following, heuristically chosen single set of parameters of our algorithm $m=0.05$, $p=10$, and $k=25$, which are the mixing factor of segments connection criteria, segments length threshold, and the number of knots.

\subsection{Datasets}
To evaluate the proposed cable tracking method (DLOFTBs), we conducted several experiments, which show the performance of the proposed algorithm on 4 datasets: 
\subsubsection{RGB real}
\label{sec:rgb_real}
7 sequences of RGB images ($\approx900$ frames in total), collected with the Intel RealSense D435 camera, of a single cable being manipulated by two UR3 manipulators. 

\subsubsection{RGBD real}
\label{sec:rgbd_real}
10 sequences of RGBD images ($\approx2500$ frames in total), collected with the Kinect Azure, of the single cable being manipulated by a human.

\subsubsection{RGBD artificial}
\label{sec:rgbd_art}
5 sequences of the artificially created RGBD images ($\approx1400$ frames in total), generated from a reference curve evolving in time. This dataset allows us to compare directly to the reference curve instead to mask.

\subsubsection{Ariadne+}
\label{sec:ariadne}
The test set, taken from~\cite{ariadne_plus}, consists of 62 images of multiple cables. We enriched this dataset with manual annotations of the cable shapes to facilitate direct comparison between curve shapes.

\subsection{Performance criteria}

Assessing the quality of the DLO shape tracking is not a trivial task~\cite{shape_similarity}, especially when the only ground truth data available is the mask of the DLO (datasets 1 and 2).
For dataset 1 and dataset 2, we use two Mean Minimal Distance (MMD) criteria, which build upon the ideas of Modified Hausdorff Distance~\cite{mhd} and are defined by
\begin{equation}
    \mathcal{L}_1 = \MMD(\mathcal{M}, \mathcal{C}_d) \quad\text{and}\quad \mathcal{L}_2 = \MMD(\mathcal{C}_d, \mathcal{M}), 
\end{equation}
where 
\begin{equation}
    \MMD(X, Y) = \frac{1}{|X|}\sum_{x \in X} \min_{y \in Y} d(x, y),
\end{equation}
where $d(x, y)$ is a Euclidean distance between $x$ and $y$, $\mathcal{M}$ is a set of pixels belonging to a mask, while $\mathcal{C}_d$ is a set of points on the predicted curve $\mathcal{C}$.

In turn, for dataset 3 and dataset 4 we have access to the mathematical curve representing the reference shape $\mathcal{C}_r$. Therefore, we can formulate a much more accurate measure of the performance, which builds upon the Fr\'echet distance ~\cite{frechet}, and is defined by
\begin{equation}
    \mathcal{L}_3(\mathcal{C}_d, \mathcal{C}_{r_d}) = \frac{F(\mathcal{C}_d, \mathcal{C}_{r_d}) + F(\mathcal{C}_{r_d}, \mathcal{C}_d)}{2},
\end{equation}
where $\mathcal{C}_{r_d}$ is a discretized version of the reference path, and where
\begin{equation}
\begin{split}
    F(X, Y) = \frac{1}{|X|}\sum_{i = 0}^{|X| - 1} \min_{w \in [0; 1]} d(&X(i), (1-w) Y(k(i)) \\
    & + w Y(k(i)+1)), 
\end{split}
\end{equation}
where $k(i)$ satisfies $D_Y(k(i)) \leq D_X(i) \leq D_Y(k(i)+1)$ and is monotonically non-decreasing, where $D_X(i)$ is a normalized distance along $X$ curve at $i$-th discretization point. This function allows for fair alignment of curves despite possible differences in parameterization.

\begin{figure}[h]
    \centering
    \includegraphics[width=\linewidth]{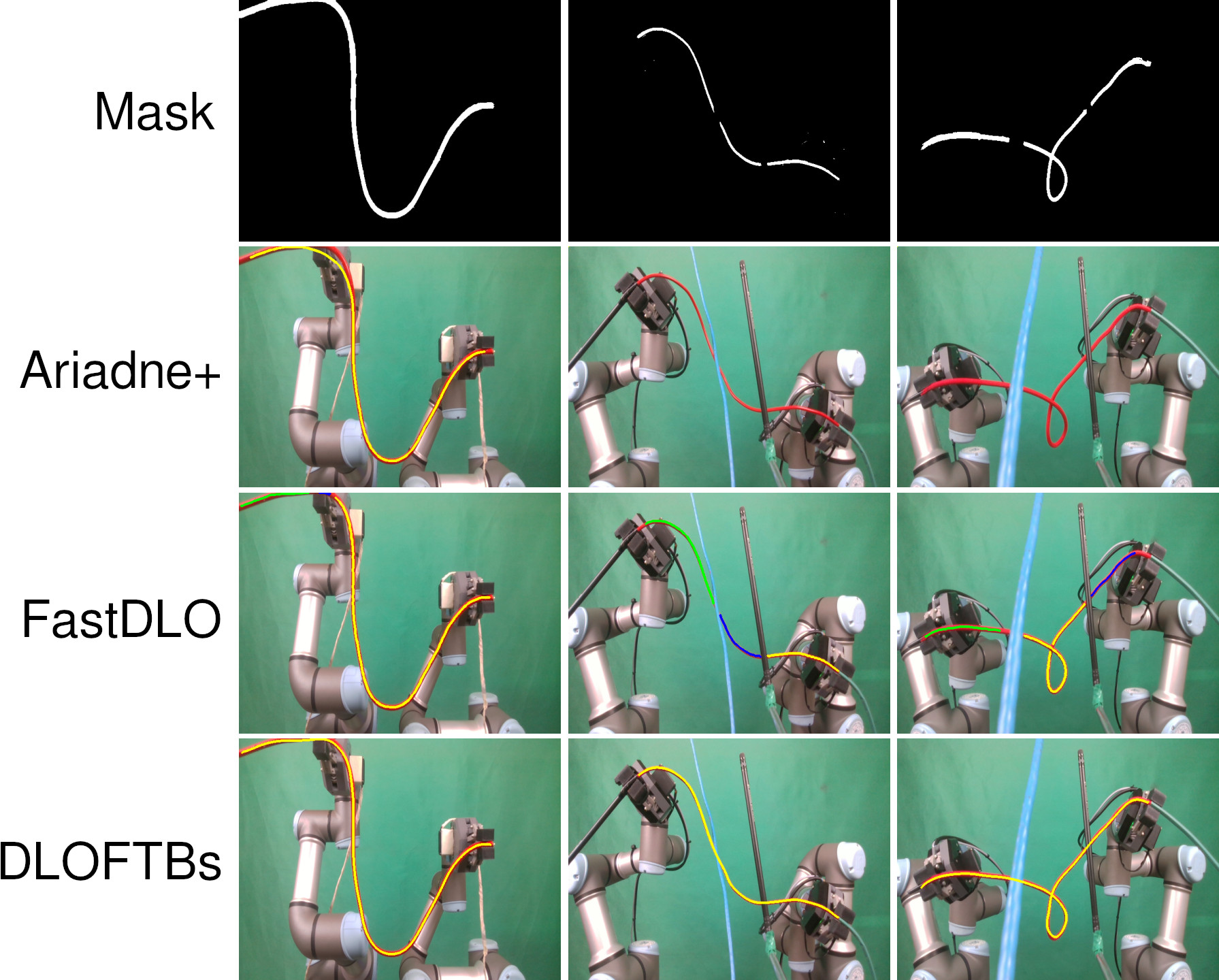}
    \caption{Comparison of 2D DLO tracking algorithms. Only our approach was able to handle occlusions and self-intersections of the tracked cable.}
    \label{fig:2d_tracking}
    \vspace{-0.15cm}
\end{figure}

\subsection{2D videos of a single cable}

\begin{table*}[!th]
\setlength{\tabcolsep}{0.5pt}
\centering
\caption{Performance of the proposed DLO tracker on several 2D videos of dual-arm manipulation.}
\begin{tabular}{cc|ccccccc}
\hline
Algorithm & Scenario & 0 & 1 & 2 & 3 & 4 & 5 & 6\\
\hline
\multirow{3}{*}{DLOFTBs} 
& $\mathcal{L}_1$ [px] & \textbf{4.82$\pm$0.05} & \textbf{6.15$\pm$0.13} & \textbf{2.45$\pm$0.09} & \textbf{2.88$\pm$0.17}  & \textbf{4.97$\pm$0.30} & \textbf{2.92$\pm$0.08} & \textbf{4.45$\pm$0.04}\\
& $\mathcal{L}_2$ [px] & 0.88$\pm$2.35 & 0.39$\pm$0.12 & 1.67$\pm$3.13 & 7.08$\pm$14.58 & 0.56$\pm$0.03 & 0.72$\pm$0.07 & 0.99$\pm$0.02\\
& Time [ms]            & \textbf{38.2$\pm$6.4}  & \textbf{34.8$\pm$4.4}  & \textbf{30.0$\pm$5.2}  & \textbf{33.2$\pm$10.0}  & \textbf{39.5$\pm$4.0}  & \textbf{26.0$\pm$2.0} & \textbf{35.1$\pm$2.0} \\
\hline
\multirow{3}{*}{Ariadne+} 
& $\mathcal{L}_1$ [px] & 8.03$\pm$4.38 & 12.45$\pm$16.80 & 30.01$\pm$32.57 & 29.55$\pm$0.64  & 54.85$\pm$71.67 & 117.60$\pm$38.23 & --\\
& $\mathcal{L}_2$ [px] & 0.51$\pm$0.8 & 1.99$\pm$4.71 & 1.22$\pm$3.18 & 47.59$\pm$1.55 & 0.74$\pm$0.23 & 2.63$\pm$6.60 & --\\
& Time [ms]            & 973.9$\pm$35.3  & 956.0$\pm$61.3  & 974.2$\pm$30.5  & 935.5$\pm$37.0  & 962.5$\pm$31.0  & 923.3$\pm$30.9 & --\\
\hline
\multirow{3}{*}{FastDLO} 
& $\mathcal{L}_1$ [px] & 9.17$\pm$12.8 & 8.98$\pm$10.32 & 29.3$\pm$33.2 & 29.1$\pm$47.5  & 177.6$\pm$16.6 & 173.7$\pm$6.5 & 171$\pm$13\\
& $\mathcal{L}_2$ [px] & \textbf{0.36$\pm$0.01} & \textbf{0.37$\pm$0.01} & \textbf{0.36$\pm$0.01} & \textbf{0.36$\pm$0.02} & \textbf{0.37$\pm$0.01} & \textbf{0.36$\pm$0.01} & \textbf{0.36$\pm$0.02}\\
& Time [ms]            & 86.3$\pm$6.3  & 97.6$\pm$10.2  & 62.2$\pm$6.5  & 64.9$\pm$9.5  & 96.4$\pm$9.1  & 60.8$\pm$9.0 & 88.3$\pm$3.6 \\
\hline

& Frames & 
\includegraphics[width=\realddwidth\textwidth,valign=c]{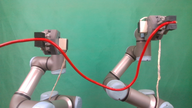} &
\includegraphics[width=\realddwidth\textwidth,valign=c]{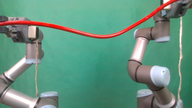} &
\includegraphics[width=\realddwidth\textwidth,valign=c]{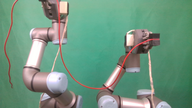} &
\includegraphics[width=\realddwidth\textwidth,valign=c]{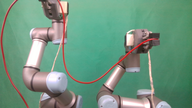} & \includegraphics[width=\realddwidth\textwidth,valign=c]{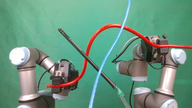} & \includegraphics[width=\realddwidth\textwidth,valign=c]{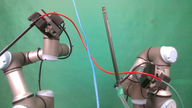} &
\includegraphics[width=\realddwidth\textwidth,valign=c]{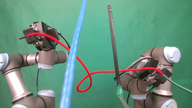} \\
\end{tabular}
\label{tab:2d_L1_L2}
\vspace{-0.1cm}
\end{table*}

In the first stage of the experimental evaluation, we evaluated DLOFTBs on \textit{RGB real} dataset (see Section~\ref{sec:rgb_real}) , with masks generated using hue-based segmentation, and compared it with Ariadne+~\cite{ariadne_plus} and FastDLO~\cite{fastdlo} learned approaches. We used cables with different widths and lengths and tested all algorithms on challenging setups, shown in Table~\ref{tab:2d_L1_L2}, including self-intersection (\textit{scenario 6}) and occlusions (\textit{scenarios 2-6}). 
Obtained results show that the proposed algorithm achieves the most stable behavior and outperforms all baselines in terms of criterion $\mathcal{L}_1$ and processing time, and achieves similar results in terms of criterion $\mathcal{L}_2$. Huge values of $\mathcal{L}_1$ for the baselines indicate, that, unlike DLOFTBs, they are unable to cover the whole cable with the predicted spline (extreme case is Ariadne+ which was unable to generate any curve for \textit{scenario 6}). Whereas, small values of $\mathcal{L}_2$ for almost all cases, ensures that predicted splines do not cover empty areas. Our method achieves a relatively big value of $\mathcal{L}_2$ only for \textit{scenario 3}, in which large parts of the cable are outside the camera's field of view, therefore, even reasonable and plausible curves generated by the proposed method results in the growth of $\mathcal{L}_2$. The behavior of the algorithms for some sample challenging frames is presented in Figure~\ref{fig:2d_tracking}. Even though both baselines were provided with a very clean mask of the tracked cable, they were unable to handle occlusions and self-intersections, whereas DLOFTBs handled them perfectly.


\begin{table*}[!b]
\setlength{\tabcolsep}{1pt}
\centering
\caption{Comparison of the DLOFTBs with CDCPD2 algorithm on several 3D videos of a human manipulating the cable.}
\begin{tabular}{cc|cccccccccc}
\hline
\multicolumn{2}{c}{Scenario} & 0 & 1 & 2 & 3 & 4 & 5 & 6 & 7 & 8 & 9\\
\hline
\multirow{3}{*}{CDCPD2~\cite{cdcpd2}} 
 & $\mathcal{L}_1$ & 6.26 & 7.1  & 6.21 & 5.73 & 15.6 & 5.49 & 6.03 & 5.79 & 19.0 & 11.19\\
 & $\mathcal{L}_2$ & 2.73 & 2.85 & 6.55 & 3.87 & 16.6 & 2.83 & 2.76 & 2.5  & 12.3 & 3.79\\
 & Time [ms]       & 81   & 110  & 125  & 120  & 87   & 85   & 91   & 92   & 92   & 103\\
\multirow{3}{*}{DLOFTBs} 
 & $\mathcal{L}_1$ & \textbf{5.05} & \textbf{5.31} & \textbf{4.41} & \textbf{4.5}  & \textbf{4.07} & \textbf{3.79} & \textbf{4.69} & \textbf{4.51} & \textbf{4.79} & \textbf{5.23}\\
 & $\mathcal{L}_2$ & \textbf{1.08} & \textbf{1.04} & \textbf{5.75} & \textbf{2.71} & \textbf{2.08} & \textbf{0.88} & \textbf{1.21} & \textbf{1.03} & \textbf{1.58} & \textbf{0.68}\\
 & Time [ms]       & \textbf{26}   & \textbf{36}   & \textbf{38}   & \textbf{33}   & \textbf{29}   & \textbf{30}   & \textbf{19}   & \textbf{18}   & \textbf{23}   & \textbf{26}\\
 \multicolumn{2}{c|}{Sample frame} &
 \includegraphics[width=\realtdwidth\textwidth,valign=c]{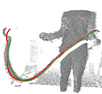} &
 \includegraphics[width=\realtdwidth\textwidth,valign=c]{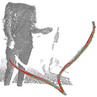} & 
 \includegraphics[width=\realtdwidth\textwidth,valign=c]{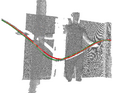} & 
 \includegraphics[width=\realtdwidth\textwidth,valign=c]{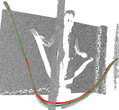} & 
 \includegraphics[width=\realtdwidth\textwidth,valign=c]{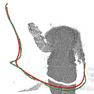} & 
 \includegraphics[width=\realtdwidth\textwidth,valign=c]{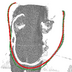} & 
 \includegraphics[width=\realtdwidth\textwidth,valign=c]{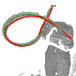} & 
 \includegraphics[width=\realtdwidth\textwidth,valign=c]{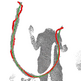} &
 \includegraphics[width=\realtdwidth\textwidth,valign=c]{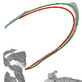} &
 \includegraphics[width=\realtdwidth\textwidth,valign=c]{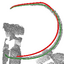}\\
\hline
\end{tabular}
\label{tab:3d_L1_L2}
\end{table*}

\subsection{2D masks of multiple cables}

\begin{figure}[t]
    \centering
    \includegraphics[width=\linewidth]{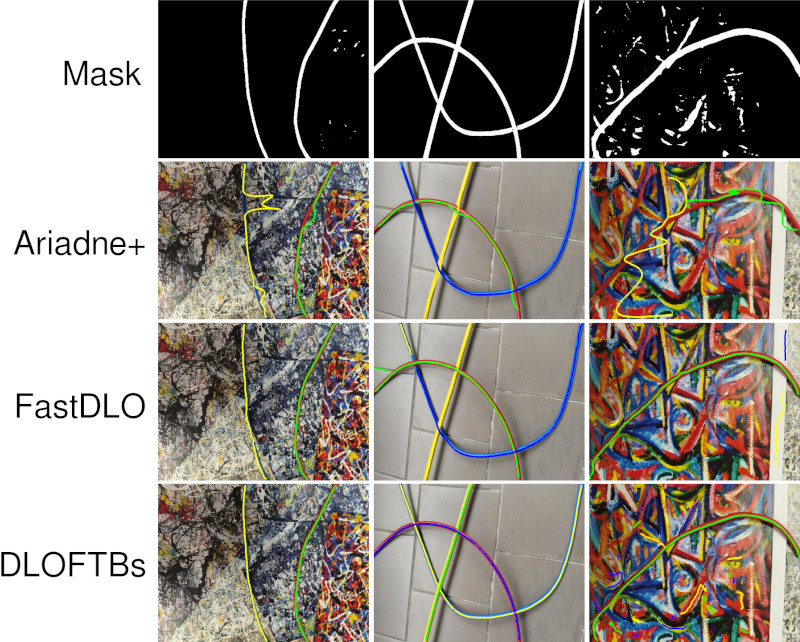}
    \caption{Comparison of multiple DLOs tracking methods on Ariadne+ dataset.}
    \label{fig:multiple_cables}
    \vspace{-0.5cm}
\end{figure}

In this experiment, we evaluated the ability of DLOFTBs to identify multiple cables at once on the masked image and compared directly with Ariadne+ and FastDLO algorithms~\cite{ariadne_plus, fastdlo} on the augmented version of the Ariadne+ test set (Section~\ref{sec:ariadne}) segmented using DeepLabV3+ network~\ref{tab:ariadne} for all algorithms. The result of this comparison can be found in Table~\ref{tab:ariadne}. We outperformed Ariadne+ and FastDLO in terms of algorithm execution time, and the accuracy of the DLO shape reconstruction, and scored second in the number of wrongly identified DLOs. The relatively high number of redundant curves fitted by DLOFTBs is a result of extremely noisy masks generated by DeepLabV3+ (see 3rd column of Figure~\ref{fig:multiple_cables}). In Figure~\ref{fig:multiple_cables} we present a qualitative analysis of algorithms behavior on 3 challenging images. Ariande+ has severe problems with handling complex backgrounds and bends at the intersections of cables, while FastDLO cannot solve the intersection in the 2nd image properly and produces a wavy shape for the left cable in the 1st image. In turn, DLOFTBs generates the most accurate solutions for the first two images, however, if the mask is very noisy (3rd image) it fits curves into linear false positives regions of the mask.


\begin{table}[htbp]
\setlength{\tabcolsep}{3pt}
\centering
\caption{Comparison of DLOFTBs and Ariadne+ algorithms on multiple cable detection benchmarks.}
\vspace{-0.1cm}
\begin{tabular}{c|cccc}
\hline
Algorithm & $\mathcal{L}_3$ & \# missing & \# redundant & Time [ms]\\
\hline
Ariadne+ & 45.06 & 9 & \textbf{16} & 421.3\\
FastDLO & 51.55 & \textbf{3} & 75 & 64.3\\
DLOFTBs & \textbf{27.17} & \textbf{3} & 33 & \textbf{39.2}\\
\hline
\end{tabular}
\label{tab:ariadne}
\vspace{-0.25cm}
\end{table}

\subsection{3D video sequences}
To accurately compare the proposed approach with another State-of-the-Art method, we made our algorithm work with 3D data, for which the State-of-the-Art CDCPD2 algorithm~\cite{cdcpd2} was designed.


\subsubsection{Real data}

\begin{figure*}[t]
    \centering
    \includegraphics[width=\linewidth]{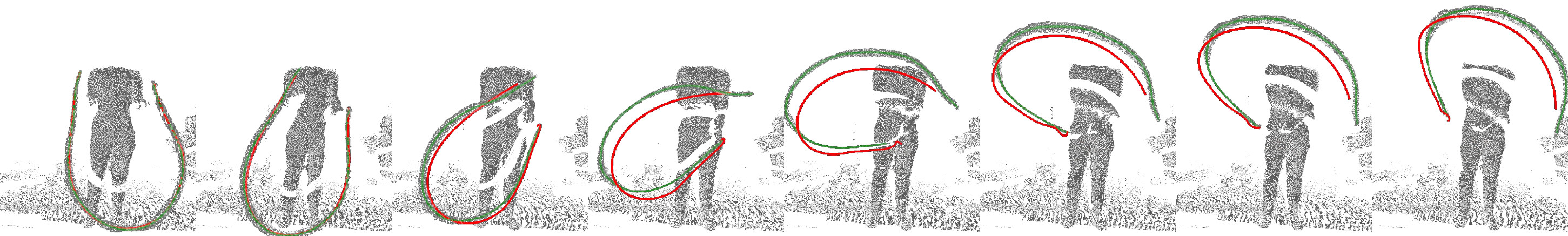}
    \caption{DLOFTBs (green) and CDCPD2 algorithm (red) on real-data 3D cable tracking sequence. With the fast movement of the cable and insufficiently large cable model updates, CDCPD2 could not track the moving cable. In contrast, our approach accurately tracks the cable shape throughout the sequence.}
    \label{fig:seq9}
    \vspace{-0.25cm}
\end{figure*}

In our experiments, to accurately compare the precision of shape tracking, we adjusted the video frame rate to enable each algorithm to process it at its own pace and reported the times needed to process a single frame.
Furthermore, because the performance of the CDCPD2 method is strongly related to the cable length estimate, we tested it for several different lengths for each scenario and reported only the best result, while our algorithm required no parameter tuning.


In Table~\ref{tab:3d_L1_L2} we present mean values of the $\mathcal{L}_1$ and $\mathcal{L}_2$ errors and mean algorithm running times for 10 scenarios of cable being manipulated by a human, and sample frames for each scenario from the \textit{RGBD real} dataset (see Section~\ref{sec:rgbd_real}).
DLOFTB achieves lower errors than CDCPD2 for all considered scenarios and criteria, and its running times are about 3 times shorter.
While for many scenarios values of the criteria are rather comparable, there are some cases where the proposed approach outperforms the CDCPD2 by a large margin (scenarios 4, 8, 9). In these cases, the CDCPD2 algorithm lost the track of the cable shape due to the fast movements of the cable (scenarios 8 and 9) or the complexity of the initial shape (scenario 4). 

In Figure~\ref{fig:seq9} we present a sample tracking sequence, in which our algorithm is able to keep track of the cable movements and deformations. At the same time, for CDCPD2 the changes are too significant to be possible to follow. Moreover, for the last images in the sequence, CDCPD2 produces a wavy shape, which does not reflect the actual cable shape but does not increase the performance error measures significantly.

\subsubsection{Artificial data}
To expose the aforementioned types of errors and accurately measure the quality of tracking, we need to utilize the $\mathcal{L}_3$ criterion.
To do so, we used \textit{RGBD artificial} dataset (see Section~\ref{sec:rgbd_art}), which also includes challenging cases like high cable curvature (\textit{scenarios 0, 1}), self-intersections (\textit{scenarios 1, 2}) and rapid cable moves (\textit{scenarios 3, 4}).

\begin{figure}[h]
    \centering
    \includegraphics[width=\linewidth]{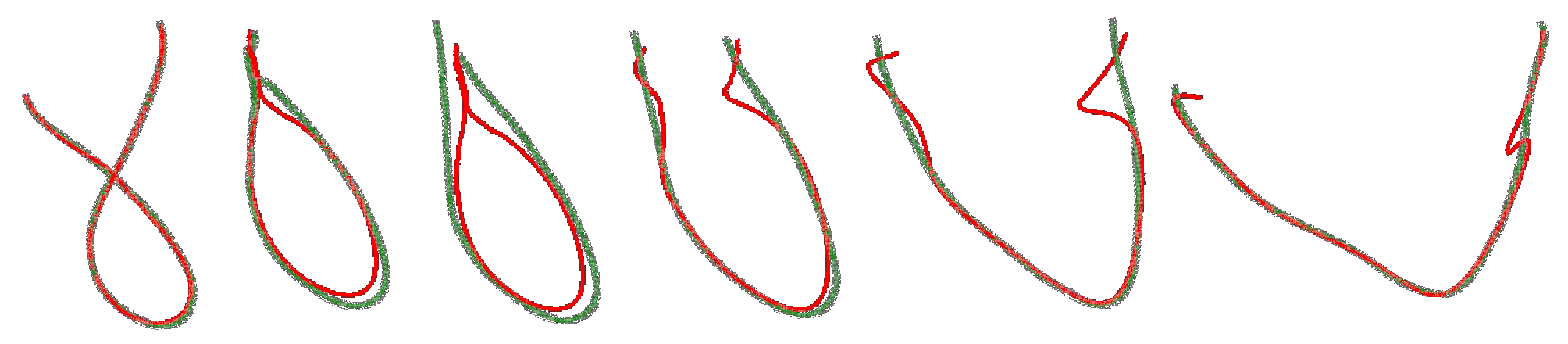}
    \caption{DLOFTBs (green) and CDCPD2 algorithm (red) on artificially generated data of 3D cable tracking sequence. In the complex case of deforming the cable from the self-intersection into the basic shape, our approach accurately tracks the deformation throughout the whole sequence. In contrast, the CDCPD2 started with accurate tracking but degraded significantly after the last moment when the cable was self-intersecting.}
    \label{fig:seq_spline}
    \vspace{-0.25cm}
\end{figure}

The results of this comparison are presented in Table~\ref{tab:3d_L3}. 
Also, in this experiment, our proposed approach outperforms the CDCPD2 algorithm. However, the use of the more accurate criterion emphasized the differences between compared methods. DLOFTB achieves mean $\mathcal{L}_3$ values that are from 6 to 20 times smaller than those achieved by CDCPD2.
Minimal mean values of $\mathcal{L}_3$ show that our approach is, on average, more accurate than the best predictions made by the CDCPD2 in 4 out of 5 scenarios. Moreover, maximal mean values show that throughout 3 out of 5 scenarios, DLOFTBs does not produce any significantly wrong measurements (max mean $\mathcal{L}_3 > 10$), while CDCPD2 does so for all scenarios. 
In Figure~\ref{fig:seq_spline} we present a part of the \textit{scenario 2} in which cable was recovering from the self-intersection. Our algorithm was able to accurately track the cable throughout the whole process. In contrast, the CDCPD2 crushed when the cable was occluding itself a moment before the untangling and lost track for the rest of the sequence.

\begin{table}[!b]
\setlength{\tabcolsep}{2.5pt}
\centering
\caption{Comparison of DLOFTBs and CDCPD2 algorithms on several artificially generated 3D cable manipulation scenarios.}
\begin{tabular}{cc|ccccc}
\hline
\multicolumn{2}{c}{Scenario} & 0 & 1 & 2 & 3 & 4\\
\hline
\multirow{3}{*}{CDCPD2} 
 & mean $\mathcal{L}_3$     & 7.94 & 10.62 & 26.8 & 23.9 & 30.4\\
 & max mean $\mathcal{L}_3$ & 128  & 75  & 198  & 82  & 100\\
 & min mean $\mathcal{L}_3$ & 1.96 & 2.96 & 2.02 & 2.82 & 2.48\\
\multirow{3}{*}{DLOFTBs} 
 & mean $\mathcal{L}_3$     & \textbf{1.59} & \textbf{4.22} & \textbf{1.99} & \textbf{2.42} & \textbf{2.47}\\
 & max mean $\mathcal{L}_3$ & \textbf{2.6} & \textbf{38.1}  & \textbf{47.7} & \textbf{5.75} & \textbf{6.91}\\
 & min mean $\mathcal{L}_3$ & \textbf{1.23} & \textbf{1.57} & \textbf{1.1} & \textbf{1.48} & \textbf{1.34}\\
 
  \multicolumn{2}{c|}{Sample frame} &
 \includegraphics[width=\arttdwidth\textwidth,valign=c]{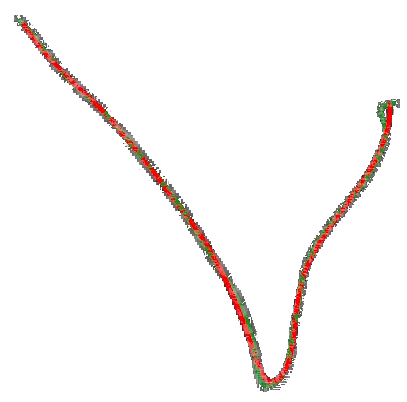} &
 \includegraphics[width=\arttdwidth\textwidth,valign=c]{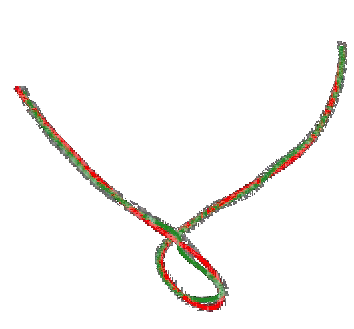} & 
 \includegraphics[width=\arttdwidth\textwidth,valign=c]{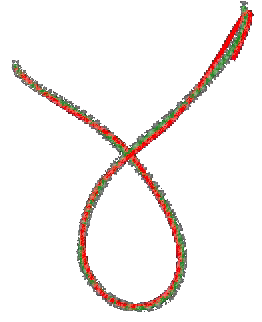} & 
 \includegraphics[width=\arttdwidth\textwidth,valign=c]{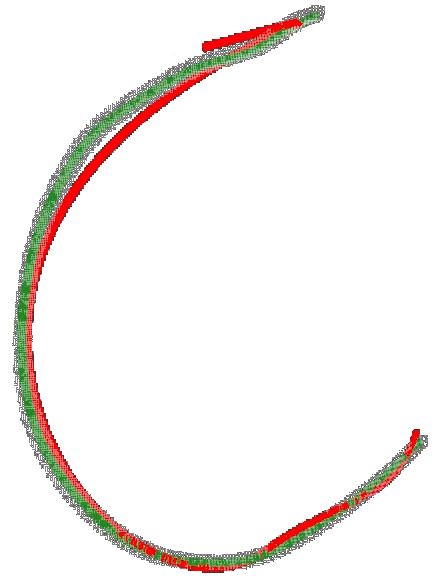} & 
 \includegraphics[width=\arttdwidth\textwidth,valign=c]{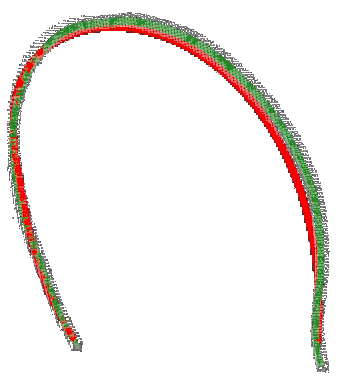} \\
\hline
\end{tabular}
\label{tab:3d_L3}
\end{table}

 

\section{Conclusions}
This paper proposes a novel approach to DLO tracking on 2D and 3D images and videos called DLOFTBs.
Using a segmented mask of the cable, we can precisely fit a B-spline representation of its shape within tens of milliseconds. The experimental analysis showed that DLOFTB is accurate and can handle tedious cases like occlusions, self-intersections, or even multiple DLOs at one time.
Moreover, it outperforms the State-of-the-Art DLO tracking algorithms CDCPD2~\cite{cdcpd2}, Ariadne+~\cite{ariadne_plus}, and FastDLO~\cite{fastdlo} in all considered scenarios both in terms of the quality of tracking, identification of multiple cables, and algorithm running time. \rev{Moreover, the proposed solution is able to solve all aforementioned problems with a single set of parameters and does not require any training. Thus it does not depend on the training data, and, unlike the CDCPD2, does not need any prior information about the DLO.}

Our method was extensively tested against algorithmic and learned methods. The weakness of the approaches that utilize learning is that they have substantial problems with generalization and are not working with the data outside the training set distribution. Our approach is not suffering from this issue, so it is better suitable for robotics. We claim that there is still some space for non-deep-learning approaches, which are better in generalization and are fully explainable. 

\addtolength{\textheight}{-10cm}   






\bibliographystyle{IEEEtran}
\bibliography{IEEEexample}


\end{document}